\documentclass{article}

\usepackage[preprint]{neurips_2026}

\usepackage[utf8]{inputenc} 
\usepackage[T1]{fontenc}    
\usepackage{hyperref}       
\usepackage{url}            
\usepackage{booktabs}       
\usepackage{amsfonts}       
\usepackage{nicefrac}       
\usepackage{microtype}      
\usepackage{xcolor}         
\usepackage{amsmath}        
\usepackage{amssymb}        
\usepackage{graphicx}       
\usepackage{subcaption}     
\usepackage{enumitem}       
\usepackage{algorithm}      
\usepackage{algpseudocode}  
\usepackage{algpseudocode}
\usepackage{placeins}       
\usepackage{float}          

\graphicspath{{Figures/}{figures/}{figs/}}

\newcommand{\maeOurs}{\textbf{3.57}}
\newcommand{\rmseOurs}{\textbf{11.34}}
\newcommand{\ssimOurs}{0.852}
\newcommand{\edgeOurs}{\textbf{0.719}}

\title{A 3D Isovist World Model: Revealing a City's Unseen
       Geometry and Its Emergent Cross-City Signature}

\author{%
  Xuhui Lin \\
  Department of Geography\\
  University College London, UK
  \And
  Stephen Law \\
  Department of Geography\\
  University College London, UK
  \And
  Nanjiang Chen \\
  School of Project Management, Faculty of Engineering\\
  The University of Sydney, AU
  \AND
  Kunyao Li \\
  School of Engineering\\
  Cardiff University, UK
  \And
  Tao Yang\thanks{Corresponding author.} \\
  School of Architecture\\
  Tsinghua University, Beijing, CN
}

\begin{document}

\maketitle

\begin{abstract}
Embodied agents that navigate cities rely on world models that predict how their
surroundings will change as they move. But for navigation, what matters is not what
the buildings look like; it is where the agent can go. Most world models nonetheless
predict appearance, learning how a scene looks rather than the space an agent can
move through. Those that do target geometry, such as bird's-eye-view occupancy grids,
flatten the three-dimensional environment onto a ground plane, discarding the
above-ground and multi-level structure that shapes real navigation. What is missing
is a predictive target that captures the navigable geometry an agent actually
traverses, without photometric entanglement and without collapsing the third
dimension. Our key idea is to model the open volume between buildings, the
\emph{negative space}, encoded as a 3D \emph{isovist}: a spherical visibility-depth
map recording the distance to the nearest surface in every direction. We introduce
an embodied world model that predicts the next isovist from a short history of past
isovists and a movement action. The prediction is formulated as a depth
\emph{residual} so the decoder inherits sharp building edges, trained with
\emph{self-rollout scheduled sampling} to keep corrupted context on the geometry
manifold, and equipped with a \emph{persistent latent bird's-eye-view spatial map}
for cross-path consistency. Our central finding is emergent and unexpected: a single
\emph{city-blind} model trained on Manhattan and Paris develops a \emph{cross-city
spatial signature}, with city identity linearly decodable from its temporal latents
far above single-frame baselines, so the signature lives in the learned dynamics
rather than in appearance. The representation is lightweight, interpretable, and
reproducible, offering a geometric substrate for spatial reasoning in embodied AI,
robotics, and urban analysis, released with an open dataset and pipeline.
\end{abstract}

\vspace{0.4em}
\noindent\textbf{Keywords:} Isovist, World Model, Embodied Navigation, Negative
Space, Urban Morphology.

\section{Introduction}
\label{sec:intro}

Consider an embodied agent moving along a city street. At every step, what it can
perceive is not the buildings themselves but the \emph{space between} them: the
open, traversable volume its body occupies, shaped by surrounding facades, corners,
and passages. This visibility volume --- the set of all points visible from a given
location --- is called an \textbf{isovist}~\citep{benedikt1979}. Formalized in
architectural theory~\citep{hillier1984} and spatial
cognition~\citep{turner2001, wiener2007isovist}, isovists have long served to
analyse how humans experience and navigate urban space. In three dimensions, an
isovist is compactly encoded as a spherical depth map
$D \in \mathbb{R}^{H \times W}$ indexed by polar angle $\theta$ and azimuth
$\varphi$, where each cell records the distance to the nearest surface in that
direction. Each step of motion produces a new isovist; the sequence of isovists
along a path is therefore a natural state representation for a navigating agent ---
and, we argue, the natural perceptual substrate for an \emph{embodied} world model
of the city.

\begin{figure}[t]
  \centering
  \includegraphics[width=\linewidth]{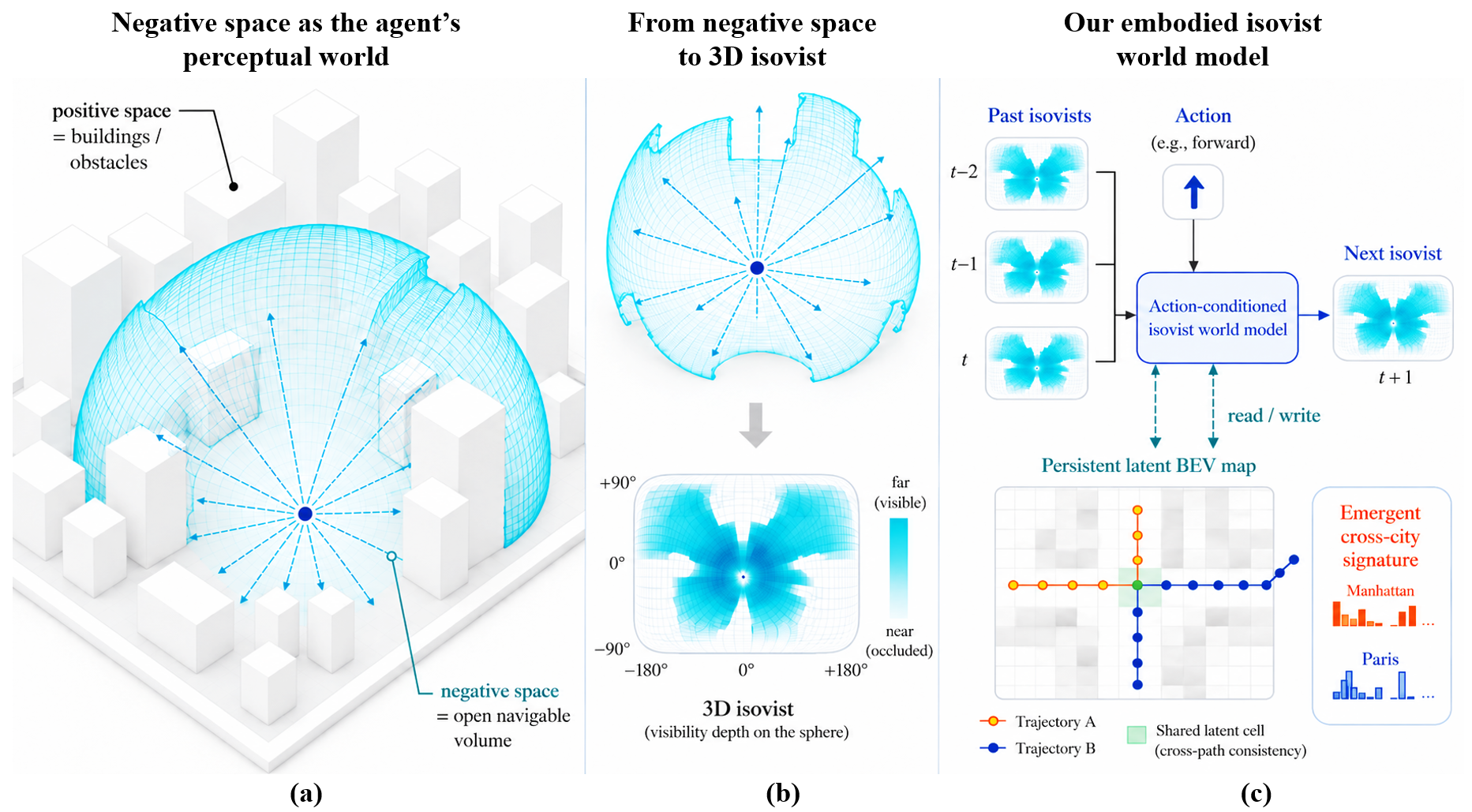}
  \caption{\textbf{Negative space as the agent's perceptual world.} (a) As an
  embodied agent moves through a city, what it perceives is not the buildings
  (positive space) but the open, navigable volume between them (negative space), the
  visibility volume bounded by surrounding facades. (b) This volume is encoded as a
  3D isovist: rays cast from the agent to the nearest surface in every direction
  yield a spherical visibility-depth map over azimuth
  $\varphi \in [-180^\circ, 180^\circ]$ and elevation
  $\theta \in [-30^\circ, 90^\circ]$, where each cell records distance to the
  nearest surface (lighter $=$ farther/visible, darker $=$ nearer/occluded). (c) An
  action-conditioned world model predicts the next isovist from a short history of
  past isovists and a movement action, and reads from / writes to a persistent
  latent bird's-eye-view (BEV) spatial map keyed by world coordinates; independent
  trajectories crossing a shared cell read and write the same memory, a mechanism
  for cross-path consistency. Trained city-blind on Manhattan and Paris, the model's
  latents carry an emergent cross-city signature (right).}
  \label{fig:framework}
\end{figure}

Despite this, isovists have never, to our knowledge, been used as the predictive
state of a world model. We argue this is a significant gap, and that closing it
changes what the model is forced to represent. From the perspective of a robot or
autonomous agent navigating a city, the central question is not ``what do the
buildings look like?'' but ``what space is available to me, and how will it change
if I move?'' The isovist answers this directly. It is \emph{perception-aligned},
encoding exactly the geometry a range sensor measures; it is
\emph{action-conditioned} by construction, since each movement produces a new
visibility volume; and it is \emph{reconstructive}, since isovists accumulated from
multiple viewpoints collectively reveal the surrounding building surfaces, with
negative space giving rise to positive-space geometry. We refer to this framing as
world modelling in \textbf{negative space} (Figure~\ref{fig:framework}).

This representational choice is the first thing that distinguishes our work, and it
is worth stating precisely what it buys an embodied agent. Appearance-first world
models predict future RGB frames~\citep{agarwal2025cosmos, wan2025,
wang2024drivedreamer} and so must carry photometric variation, such as lighting,
shadow, and surface colour, that is irrelevant to where an agent can move; geometry
is entangled with appearance, and geometric consistency becomes a secondary concern.
Bird's-eye-view occupancy models~\citep{zheng2024occworld} are geometry-pure but
collapse the full three-dimensional structure of the environment onto a flat ground
plane, discarding the vertical and above-ground geometry that shapes real
navigation, such as overpasses, stacked facades, and multi-level plazas and
frontage. The spherical isovist sits between these: it is metric geometry with no
photometric channel to entangle, and because it records depth in every direction
around the agent rather than a flattened footprint, it preserves the above-ground
and multi-level structure a ground-plane grid loses. In this sense the isovist
reveals what is otherwise hidden. It exposes the navigable geometry of the space
itself, independent of how that space happens to be lit or coloured, which is
precisely the structure that supports how embodied agents move through cities, on
the ground and above it. Implicit scene representations such as
NeRF~\citep{mildenhall2020nerf} and 3D Gaussian Splatting~\citep{kerbl20233dgs}
reconstruct seen scenes at high fidelity but are scene-specific and not designed for
action-conditioned generation across novel trajectories. Our model instead starts
from geometry directly: given a sequence of past isovists and a movement action,
predict the next isovist, replacing abstract latent features and photometric input
with an interpretable, sensor-proximal target.

A world model that merely predicts the next frame, however, has no memory of
\emph{where} it has been. When two independently sampled paths cross the same
intersection, a frame-only model can predict mutually contradictory geometry at the
shared location, because nothing ties the two predictions together. To address this,
we augment the predictor with a \textbf{persistent latent BEV spatial map}: an
explicit grid of latent features over world coordinates that the model reads from at
its current position and writes to after each observation. Because the map is keyed
by absolute world position rather than by trajectory, two paths crossing a cell read
and write the \emph{same} memory, providing a mechanism for cross-path geometric
consistency. This explicit, writable map is the structural contrast with the closest
prior work, Gornet and Thomson's emergent \emph{implicit} cognitive map for
localization in a small, fixed environment~\citep{gornet2024cognitive}; ours is
explicit, persistent, and writable, intended as a generation substrate over real
cities (Section~\ref{sec:lit}).

Training an autoregressive isovist predictor exposes a familiar failure mode: the
teacher-forcing gap. A model trained only on ground-truth context meets its own
imperfect predictions at inference, and errors compound over a
rollout~\citep{bengio2015scheduled}. The natural fix, corrupting the context during
training, has a subtlety specific to our representation: Gaussian-blur corruption of
a depth map produces something that is not a valid isovist, because it smooths
across the sharp depth discontinuities at building edges, a pattern that never
occurs in real geometry. We instead use \textbf{self-rollout scheduled sampling}: a
no-grad forward pass yields the model's own prediction, and the last context frame
is replaced by a convex blend of ground truth and that prediction on a linear
curriculum. Because the prediction is itself a valid isovist, the corrupted context
stays on the geometry manifold and better approximates the inference-time error
distribution.

We instantiate these ideas on a reproducible dataset built directly from
OpenStreetMap~\citep{haklay2008osm}. Two morphologically distinct cities (gridded
Manhattan and Haussmannian Paris) are turned into ray-cast isovist sequences along
intersection-anchored pedestrian paths, so that independently sampled paths
\emph{share} mid-route locations. We then train a single city-blind model (no city
label as input) on the combined data. The standout result is emergent: city
identity is linearly decodable from the model's temporal latents at $89.3\%$
five-fold accuracy, far above raw-pixel and single-frame-statistic probes under an
identical protocol. The model has internalized a city-distinctive spatial signature
without ever being told which city it is in. This is the same phenomenon
\citet{gornet2024cognitive} report for emergent implicit maps, observed here at the
level of city morphology and on real geometry. We are careful about what this does
and does not show: the result rests on two cities, and we disclose a
height-provenance confound (Section~\ref{sec:disc}) rather than hide it.

Based on the proposed framework, this research makes four contributions.
\begin{enumerate}[leftmargin=1.4em,itemsep=2pt,topsep=2pt]
  \item \textbf{Problem framing.} We cast embodied world modelling in
  \emph{negative space}. The task is to predict the agent's 3D isovist, a spherical
  visibility-depth map, as it moves through a city. We motivate this target as
  perception-aligned, action-conditioned, and reconstructive. Unlike RGB prediction,
  it carries no photometric entanglement; unlike flattened occupancy grids, it
  preserves the above-ground and multi-level geometry that supports navigation.
  \item \textbf{Method.} We introduce an action-conditioned autoregressive isovist
  predictor. It combines a depth-CNN and anchor-frame encoder, an
  arc-length--indexed Transformer, and a residual depth decoder. We augment it with
  a \emph{persistent latent BEV spatial map} (bilinear read, EMA or differentiable
  write) as a mechanism for cross-path consistency. We train it with self-rollout
  scheduled sampling, which keeps the corrupted rollout context on the geometry
  manifold.
  \item \textbf{Empirical headline.} We report an emergent \emph{cross-city spatial
  signature}. City identity is linearly decodable from a city-blind world model's
  latents at $89.3\% \pm 3.0\%$, beating the strongest single-frame baseline (raw
  pixels, $78.5\%$) by $\sim$11 points and trivial frame statistics ($69.4\%$) by
  $\sim$20. This is evidence that the signature is not single-frame appearance.
  \item \textbf{Dataset.} We release a reproducible OpenStreetMap-based
  isovist-sequence generator and a publicly released two-city dataset (Manhattan and
  Paris) under the Open Database License (ODbL), with a documented path to a broader
  multi-city release.
\end{enumerate}

The remainder of the paper is organized as follows. Section~\ref{sec:lit} situates
our work against the isovist and space-syntax literature, embodied world models,
emergent spatial maps, and the training and data-provenance practices we build on.
Section~\ref{sec:method} details the negative-space representation, the
dataset-generation pipeline, the model architecture, the persistent BEV spatial map,
and the training objective and curriculum. Section~\ref{sec:exp} presents our
experiments: single-step prediction quality against a strong copy-last baseline, the
emergent cross-city signature and its baseline ladder, the reconstruction of
positive-space geometry, and a preliminary spatial-map consistency ablation.
Section~\ref{sec:disc} discusses what the signature does and does not imply, its
relation to emergent cognitive maps, and honest limitations, before
Section~\ref{sec:conc} concludes.

\section{Literature Review}
\label{sec:lit}

This paper sits at the intersection of two lines of work, organized around two
questions an embodied negative-space world model raises. The first concerns what an
embodied navigation world model should predict. This positions our negative-space
isovist target against appearance-first, occupancy, and reconstruction-based
representations, and asks which representation actually reveals the geometry that
supports navigation. The second concerns whether predictive training on that target
induces spatial structure on its own. This connects our emergent cross-city
signature to work on emergent cognitive maps. We treat the isovist literature as the
foundation underlying both.

\subsection{Foundation: isovist analysis and space syntax}
\label{sec:lit-isovist}

The isovist, the set of all points visible from a location, was introduced
by~\citet{benedikt1979} to quantify spatial experience in architecture, and became
a central tool in the space-syntax program of~\citet{hillier1984}, which relates the
configuration of urban open space to patterns of human movement. A line of
subsequent work made the representation progressively more analytic and
quantitative. \citet{batty2001} treated the isovist as a field over space, deriving
geometric measures (area, perimeter, compactness, occlusivity) that classify the
visual character of a location, while \citet{turner2001} formalized isovists into
visibility graphs, enabling graph-theoretic measures of integration and connectivity
across a plan. \citet{morello2009} extended the analysis from the ground plane into
three dimensions, computing 3D isovists over voxelized urban form, and a body of
behavioural work linked isovist properties to how people perceive, remember, and
move through space~\citep{wiener2007isovist, franz2008}. What unites this tradition
is its stance toward the isovist: it is a descriptive statistic, computed from a
fully known plan to characterize a space that already exists and to explain observed
behaviour after the fact. We use the same representation but invert this stance,
treating the isovist as a predictive state forecast by an embodied agent rather than
measured from a known map. We develop this contrast, and our positioning against
world models, occupancy, and reconstruction, in the sections below.

\subsection{What to predict: embodied world models and action-conditioned
prediction}
\label{sec:lit-worldmodels}

Predicting future observations conditioned on actions is a long-standing goal in
model-based RL and embodied AI~\citep{ha2018world, hafner2019dream,
hafner2023dreamerv3}. The dominant recent paradigm is appearance-first: large-scale
systems generate future RGB video frames directly~\citep{wan2025,
agarwal2025cosmos}, and interactive environments learn controllable latent action
spaces from video~\citep{bruce2024genie}. Because the prediction target is pixels,
these models must carry photometric variation such as lighting, shadow, and surface
colour. This variation is irrelevant to where an agent can move, and geometric
consistency is at best a secondary concern. In autonomous driving,
DriveDreamer~\citep{wang2024drivedreamer} and UniSim~\citep{yang2023unisim}
synthesize future sensor observations conditioned on ego-motion, and
GenAD~\citep{law2024} scales video prediction to large web-collected driving
corpora. All three, however, likewise rely on dense RGB supervision. Bird's-eye-view
occupancy world models~\citep{zheng2024occworld} move in the geometry-pure direction
we favour, forecasting structure rather than appearance. But they collapse the full
three-dimensional environment onto a flattened ground-plane grid, discarding the
vertical and above-ground geometry that shapes real navigation. Our prediction
target sits deliberately between these failure modes: it is metric geometry with no
photometric channel to entangle, yet, being a spherical depth map over all viewing
directions, it preserves the multi-level and above-ground structure a flattened
occupancy grid loses. None of the prior systems model the agent's full visibility
volume as a first-class predictive target, and so none recover the complete
navigable geometry, on and above the ground, that an embodied agent actually
inhabits.

\subsection{Generation and reconstruction as alternative targets}
\label{sec:lit-reconstruction}

Two adjacent families predict geometry for different ends. The first is generative
3D city and scene models, such as CityDreamer~\citep{xie2024citydreamer},
SceneDreamer~\citep{chen2023scenedreamer}, and Infinite
Nature~\citep{liu2021infinitenature}. These synthesize unbounded environments
optimized for visual plausibility. They generate appearance and coarse geometry that
look convincing to a viewer, but not the metric structure an agent could navigate
by. The second is implicit reconstruction methods, such as
NeRF~\citep{mildenhall2020nerf}, 3D Gaussian Splatting~\citep{kerbl20233dgs}, neural
SDFs~\citep{ortiz2022isdf}, and neural SLAM~\citep{zhu2022niceslam}. These recover
detailed scenes from posed images at high fidelity but are fundamentally
scene-specific. They optimize per-scene parameters and neither generalize to novel
environments nor predict across trajectories. Monocular depth
estimators~\citep{bochkovskii2024depthpro, piccinelli2024unidepth} are closest to
our representation at the single-frame level but are neither sequential nor
action-conditioned. They answer, ``what is the depth here?'' rather than ``how will
the visibility volume change if I move?'' Our objective is distinct from all three:
we predict the metric visibility volume an embodied agent would measure, grounded in
real road networks and footprints rather than synthesized for appearance.

\subsection{Emergent structure: cognitive and spatial maps from prediction}
\label{sec:lit-emergent}

The work most closely related to ours is~\citet{gornet2024cognitive}, who show that
a visual predictive-coding network trained to predict the next view in a small,
fixed environment develops an emergent, implicit cognitive map whose latents support
localization. We share their central premise, that predictive training yields
spatially meaningful representations. Our cross-city signature is a real-city
instance of the same phenomenon, observed at the level of city morphology rather
than within-environment position. We differ on two axes. On memory, their map is
emergent and implicit, read out post hoc for localization in a single small
Minecraft-style world, whereas ours is an explicit, persistent, writable BEV latent
grid keyed by world coordinates and used as a generation substrate to enforce
cross-path consistency across real cities. On target, their predictions are over
appearance, ours over geometric negative space. More broadly,
neuroscience-inspired models report grid-like and place-like codes emerging from
navigation objectives in artificial agents~\citep{banino2018grid, cueva2018emergence};
we observe an analogous unsupervised emergence of a city-level code with no
localization or city-classification term in the loss. We draw this analogy narrowly,
as a shared high-level observation that navigation-prediction objectives induce
decodable spatial structure, not as a claim of mechanistic correspondence.

Taken together, these lines leave a specific opening: space syntax studies isovists
only as descriptive statistics; world models predict appearance or flattened
occupancy rather than the visibility volume; and emergent-map work shows prediction
induces structure, but on appearance targets in small, fixed worlds. No prior work
treats the isovist as the learnable predictive state of an embodied world model over
real cities. Our contribution targets the following gaps:
\begin{enumerate}[leftmargin=1.6em,itemsep=3pt,topsep=3pt]
  \item \textbf{The isovist is descriptive, never predictive.} In the space-syntax
  tradition the isovist is a statistic read off a known plan to explain observed
  movement; it has never been used as the action-conditioned target a navigating
  agent must forecast without access to the underlying scene.
  \item \textbf{World-model targets entangle appearance or flatten geometry.}
  Appearance-first models carry photometric variation irrelevant to navigation,
  while bird's-eye-view occupancy collapses the environment onto a ground plane.
  Neither recovers the full three-dimensional, above-ground visibility volume an
  embodied agent actually traverses.
  \item \textbf{Reconstruction methods are scene-specific, not generative across
  trajectories.} NeRF, Gaussian splatting, and neural SLAM recover seen scenes at
  high fidelity but optimize per-scene parameters and cannot predict the visibility
  volume for novel, unseen paths.
  \item \textbf{Emergent spatial memory is implicit and confined to small fixed
  worlds.} Prior emergent cognitive maps are read out post hoc for localization in a
  single small environment; there is no explicit, persistent, writable spatial
  substrate that enforces geometric agreement between independent paths crossing the
  same place in a real city.
\end{enumerate}

\section{Methodology}
\label{sec:method}

\subsection{Negative-space representation}
\label{sec:method-repr}

An \textbf{isovist} at observation point $\mathbf{p} \in \mathbb{R}^3$ is a
spherical depth map $D \in \mathbb{R}^{H \times W}$, where each cell $(i,j)$ records
the distance to the nearest scene surface along the ray with polar angle
$\theta_i$ and azimuth $\varphi_j$:
\begin{equation}
  D(\theta, \varphi) = \min \bigl\{r > 0 : \mathbf{p} + r \cdot
  \mathbf{d}(\theta,\varphi) \in \partial\mathcal{S} \bigr\},
  \label{eq:isovist}
\end{equation}
where $\mathbf{d}(\theta,\varphi)$ is the unit direction and $\partial\mathcal{S}$
is the building-surface boundary. We use $H=64$ rows over a polar range covering
elevation $[-30^\circ, 90^\circ]$ and $W=128$ columns over azimuth
$\varphi \in [0, 2\pi)$, rotated so $\varphi=0$ aligns with the agent's walking
direction; the representation is thus egocentric and invariant to global heading.
Depths are clamped to the observation radius $R_{\max} = 100$\,m; rays that miss all
geometry (sky, open ground) or exceed $R_{\max}$ are coded as zero.

The movement from step $t$ to $t{+}1$ is a 5-DoF action vector
\begin{equation}
  \mathbf{a} = \tfrac{1}{R_{\max}}\Bigl[
    \Delta_{\text{fwd}},\;
    \Delta_{\text{rgt}},\;
    \Delta_{\text{up}},\;
    \tfrac{R_{\max}}{\pi}\Delta_{\text{ang}},\;
    \ell
  \Bigr] \in \mathbb{R}^5,
  \label{eq:action}
\end{equation}
where $(\Delta_{\text{fwd}}, \Delta_{\text{rgt}}, \Delta_{\text{up}})$ is the
displacement in the local egocentric frame, $\Delta_{\text{ang}}$ is the heading
change via $\operatorname{atan2}$ (positive $=$ left turn), and $\ell$ is the step
length. Given a context window of $T=4$ historical depth maps and their actions
$\{(D_t, \mathbf{a}_t)\}$, the model predicts the next isovist $\hat{D}_{T+1}$.

\subsection{Dataset generation pipeline}
\label{sec:method-data}

We construct training data per city via a pipeline that turns an administrative
boundary polygon into ray-cast isovist sequences. The same procedure is applied
without modification to each city, supporting reproducibility and unbiased
cross-city comparison (Figure~\ref{fig:dataset}).

\paragraph{Source data.}
City boundaries are retrieved from OpenStreetMap via Nominatim. We use Manhattan,
New York (the full borough, area $\approx 115$\,km\textsuperscript{2}) and Paris,
France (arrondissements 1--6, area $\approx 15.5$\,km\textsuperscript{2}). The
walking road graph is fetched with \texttt{osmnx} and simplified to collapse
degree-two nodes while preserving intersection topology
($36{,}212$ nodes for Manhattan, $10{,}957$ for Paris). Building footprints are
fetched via the OSM \texttt{building} tag, retaining tag metadata for height
backfill.

\paragraph{Height backfill.}
OSM building heights are unevenly tagged. We assign a height to every building via a
deterministic priority: (i)~an explicit \texttt{height} tag (parsed to meters with
unit detection); (ii)~\texttt{building:levels} $\times\,3$\,m; (iii)~the median over
$K{=}8$ nearest neighbours with an assigned height; (iv)~a 10\,m global default. We
record per-building source attribution. Manhattan reaches $96.4\%$ \emph{direct} OSM
height coverage, with the remainder imputed by neighbour median. Paris reaches only
$65.8\%$ direct coverage ($1.3\%$ tag $+\,64.5\%$ levels), with $34.2\%$ imputed by
neighbour median; the 10\,m default is never triggered for either city. We disclose
the substantial Paris height imputation explicitly, as it bears on the
interpretation of the cross-city signature (Section~\ref{sec:disc}).

\paragraph{Extrusion and intersection-anchored bundles.}
Footprints are reprojected to local UTM, recentred, and extruded to their backfilled
heights into watertight meshes. Crucially, rather than the conventional
origin--destination shortest-path sampling (which yields paths that share endpoints
but rarely cross mid-route), we generate paths around \emph{intersection anchors}.
Per city we select $50$ anchor nodes by farthest-first sampling under
$\text{degree}\!\geq\!4$ and minimum pairwise distance $300$\,m, and at each anchor
walk outward along six (incoming, outgoing) neighbour pairs spanning straight
traversals and sharp turns, yielding $\approx$100--200\,m paths. Because up to six
paths pass through a common anchor, anchor-bundled paths \emph{share} mid-route BEV
cells. This is the property the persistent spatial map exploits and the cross-path
consistency experiment measures.

\paragraph{Ray-cast isovists and preprocessing.}
At each observation point (sampled every $10$\,m, observation height $1.6$\,m) we
cast $64\times128$ rays against the extruded building meshes and record the
first-hit distance (Eq.~\ref{eq:isovist}). Per-frame maps are passed through the
preprocessing pipeline (coordinate transform, $K{=}32$ geometrically salient anchor
detection, sparse-to-dense projection, action encoding) and assembled into
sliding-window samples with context length $4$ and rollout length $3$. We split
\emph{by path identity} (not by frame) $8\!:\!1\!:\!1$, then concatenate the two
cities into a combined dataset of $7{,}692$ train / $946$ validation / $832$ test
samples (validation: $609$ Manhattan, $337$ Paris). A per-sample city label is
stored for the cross-city probe but is \emph{never} fed to the model: training is
city-blind.

\subsection{Model architecture}
\label{sec:method-arch}

\begin{figure}[t]
  \centering
  \includegraphics[width=\linewidth]{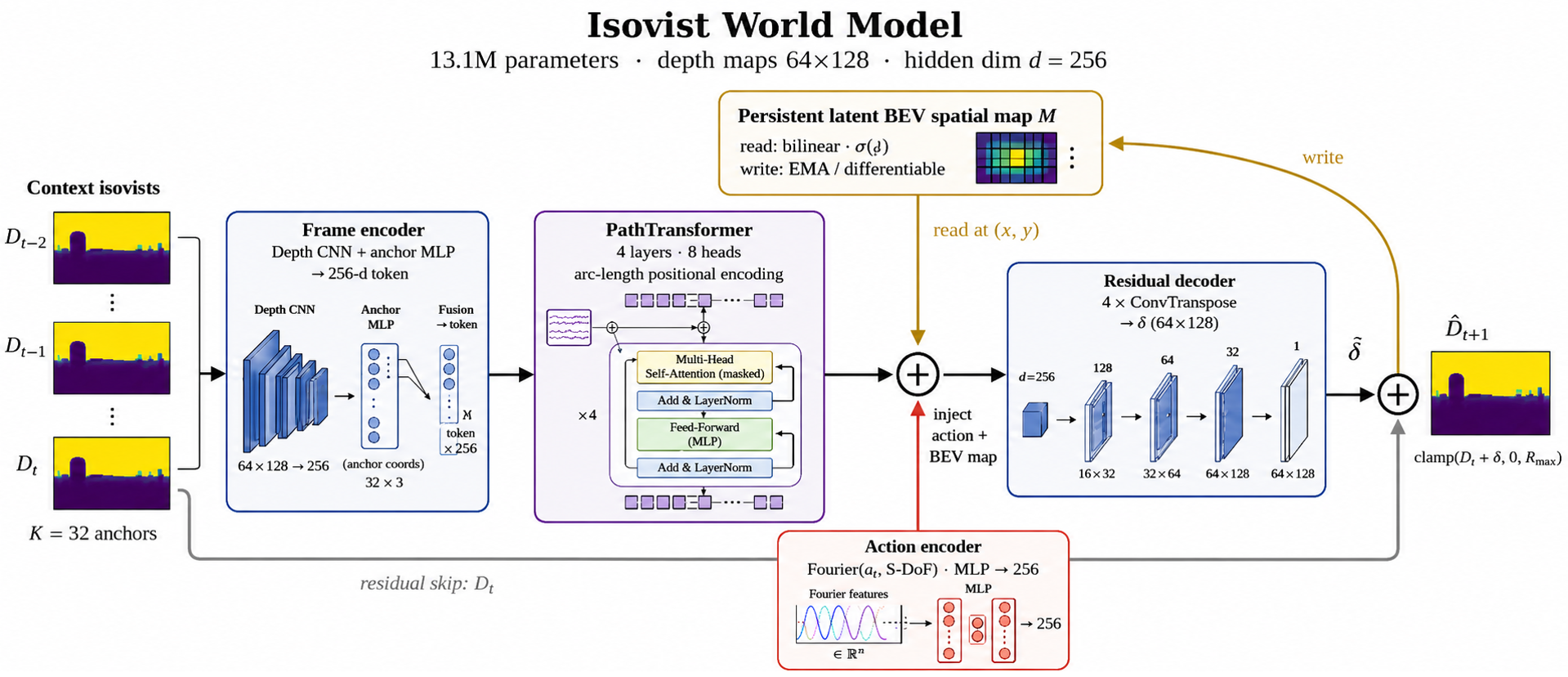}
  \caption{\textbf{Model architecture.} Each context frame is encoded by a depth CNN
  and an anchor MLP ($K{=}32$ anchors, 3-D coordinates) and fused into a 256-d
  token. An arc-length--indexed PathTransformer (4 layers, 8 heads) aggregates the
  token sequence. The Fourier-embedded 5-DoF action and the BEV-map read feature are
  injected into the context summary (the central $\oplus$). A residual decoder of
  four ConvTranspose layers produces a depth-change map $\boldsymbol{\delta}$, which
  is added to the most recent frame $D_T$ and clamped (Eq.~\ref{eq:residual}). The
  persistent BEV spatial map is read at the current world position (bilinear) and
  written back (EMA / differentiable) after the observation.}
  \label{fig:arch}
\end{figure}

The model follows an encode--aggregate--decode pipeline (Figure~\ref{fig:arch}) with
hidden dimension $d=256$ throughout. It has $13.07$M parameters ($13.14$M with the
spatial map).

\paragraph{Frame encoder.}
Each context frame $D_t \in \mathbb{R}^{64 \times 128}$ is encoded by two parallel
branches. A \emph{depth CNN} applies four stride-2 convolutions
($1\!\to\!32\!\to\!64\!\to\!128\!\to\!256$, each with BatchNorm and ReLU), an
adaptive average pool to $4\times8$, a flatten, and a linear map $8192 \to 256$ (no
final activation). An \emph{anchor encoder} consumes the $K{=}32$ geometrically
salient anchor points: an anchor MLP ($3\!\to\!64\!\to\!128$), a patch MLP
($24\!\to\!64\!\to\!128$) over the 8-neighbour patch around each anchor, a fused
linear $256\!\to\!256$ with ReLU, a max-pool over the $K$ anchors, and a global
linear $256\!\to\!256$. The two 256-d features are concatenated and projected by a
linear $512 \to 256$ into the frame token $\mathbf{f}_t \in \mathbb{R}^{256}$.

\paragraph{Temporal model (PathTransformer).}
The $T$ frame tokens are processed by a custom pre-norm
Transformer~\citep{vaswani2017attention} with $4$ layers, $8$ heads, and a GELU
feed-forward block $256\!\to\!1024\!\to\!256$. For numerical stability the attention
logits are clamped to $[-50, 50]$ and the softmax is computed in fp32. Critically,
positional encoding is \emph{not} based on integer step indices but on the
cumulative \emph{arc-length} $s_t = \tfrac{1}{R_{\max}}\sum_{k\le t}\ell_k$ of the
trajectory, embedded sinusoidally. This makes the positional signal continuous and
invariant to irregular step spacing. For example, $10$\,m steps yield much smaller
inter-frame deltas than $50$\,m steps, and arc-length encoding represents this
faithfully where index encoding cannot.

\paragraph{Action encoder.}
The action $\mathbf{a}\in\mathbb{R}^5$ is embedded with Fourier
features~\citep{tancik2020fourier}: a fixed random frequency matrix produces
$\sin/\cos$ features ($\to 16$ dims), a linear map to $64$, and an MLP
($64\!\to\!128\!\to\!256$) projects to $\mathbb{R}^{256}$. The action feature is
\emph{added} to every temporal token, conditioning the aggregation on the intended
movement. The Transformer output at the final step, $\mathbf{z}\in\mathbb{R}^{256}$,
is the context summary.

\paragraph{Residual depth decoder.}
A linear map $256 \to 8192$ is reshaped to $(256,4,8)$ and upsampled by four
ConvTranspose2d layers (kernel $4$, stride $2$,
$256\!\to\!128\!\to\!64\!\to\!32\!\to\!16$), followed by a $1\times1$ Conv2d
$16\!\to\!1$ and a $\tanh$ scaled by $R_{\max}$, producing a depth-change map
$\boldsymbol{\delta}\in[-R_{\max}, R_{\max}]$. Rather than predicting the next depth
map directly, the model predicts a \emph{residual} relative to the most recent
frame:
\begin{equation}
  \hat{D}_{T+1} = \operatorname{clamp}\!\bigl(D_T + \boldsymbol{\delta},\; 0,\; R_{\max}\bigr).
  \label{eq:residual}
\end{equation}
For small ($\sim$10\,m) steps the isovist changes incrementally: facades visible in
$D_T$ largely persist, with new surfaces appearing at the field-of-view edges.
Predicting $\boldsymbol{\delta}$ lets the decoder \emph{inherit} sharp depth
discontinuities at building edges from $D_T$ rather than regenerating them, and the
$\tanh$ gate biases the decoder toward the small changes typical of pedestrian-scale
motion ($\|\boldsymbol{\delta}\| \ll R_{\max}$), stabilizing early
training~\citep{he2016resnet}.

\subsection{Persistent latent BEV spatial map}
\label{sec:method-map}

To give the model geometric memory across paths that traverse the same area, we
maintain a 2D bird's-eye-view feature grid
$M \in \mathbb{R}^{H_M \times W_M \times d}$ ($d{=}256$) over world coordinates
$(x,y)$ at cell resolution $5$\,m, initialized to zero.

\paragraph{Read and write.}
A \emph{read} at continuous world position $(x,y)$ returns the bilinear
interpolation of the four enclosing cells. A \emph{write} of feature $f$ distributes
$f$ to the four enclosing cells with bilinear weights $w$, blending into existing
content by exponential moving average:
\begin{equation}
  M[i,j] \leftarrow (1 - \alpha w_{ij})\, M[i,j] + \alpha w_{ij}\, f,
  \qquad \alpha = 1 - \rho_{\text{ema}} = 0.3,
  \label{eq:write}
\end{equation}
with EMA decay $\rho_{\text{ema}}=0.7$; the first write to an empty cell stores the
full feature.

\paragraph{Injection.}
On each forward pass the read feature $s$ is injected into the temporal summary
through a learned projection and a sigmoid gate:
\begin{equation}
  F_{\text{cond}} = F + \sigma(g)\cdot \texttt{Linear}(s),
  \label{eq:inject}
\end{equation}
where the scalar gate $g$ is initialized to $-5$ so $\sigma(g)\approx0.007$; the
map's influence ramps up only as its contents become informative during training.

\paragraph{Differentiable scene-level training.}
A \emph{differentiable} write variant produces the new map tensor out-of-place,
preserving the autograd graph so gradients flow from a later path's loss back into
the encoder weights of earlier writers in the same scene. Independent paths through a
shared region thus read and write the same cells, and a consistency term ties their
features together at common locations:
\begin{equation}
  \mathcal{L}_{\text{cons}} = \bigl\| \text{enc}(D_t) - M[\text{pos}_t] \bigr\|_2^2,
  \label{eq:cons}
\end{equation}
producing a symmetric gradient signal that incentivizes mutually consistent geometry
where paths cross.

\subsection{Training objective and curriculum}
\label{sec:method-train}

\paragraph{Edge-weighted depth loss.}
Let $g_{ij}$ be the normalized Sobel magnitude $|G_x|+|G_y|$ of the ground-truth
depth map. We define the pixel edge weight $w_{ij} = 1 + 3\,g_{ij} \in [1,4]$,
up-weighting pixels near building edges, and a weighted log-depth error (with
$\phi(x) = \log(1 + x)$, i.e.\ \texttt{log1p}):
\begin{equation}
  \mathcal{L}_{\text{depth}} = \frac{1}{HW}\sum_{i,j} w_{ij}\,
    \bigl| \phi(\hat{D}^{ij}_{T+1}/R_{\max})
         - \phi(D^{ij}_{T+1}/R_{\max}) \bigr|.
  \label{eq:ldepth}
\end{equation}

\paragraph{Gradient loss and total.}
A Sobel-edge consistency term encourages sharp boundaries, and the total loss is
\begin{equation}
  \mathcal{L}_{\text{grad}} = \frac{1}{HW}\sum_{i,j}
    \bigl| \operatorname{Sobel}(\hat{D}_{T+1})_{ij} - \operatorname{Sobel}(D_{T+1})_{ij}\bigr|,
  \qquad
  \mathcal{L} = \mathcal{L}_{\text{depth}} + 2.0\,\mathcal{L}_{\text{grad}}.
  \label{eq:total}
\end{equation}
For multi-step rollout (up to $K=3$ steps) the per-step losses are weighted by
$1/(1+0.5k)$. We optimize with AdamW~\citep{loshchilov2019adamw} (learning rate
$10^{-4}$, weight decay $0.01$), a $2000$-step warmup then cosine decay, gradient
clipping at $1.0$, and fp16 mixed precision.

\paragraph{Self-rollout scheduled sampling.}
At each step we first run a no-grad forward pass to obtain the model's own prediction
$\hat{D}_T$, then replace the last context frame with the convex blend
\begin{equation}
  \tilde{D}_T = (1-\lambda)\,D_T + \lambda\,\hat{D}_T,
  \label{eq:srss}
\end{equation}
with $\lambda$ ramping linearly $0 \to 0.7$ over training (Algorithm~\ref{alg:srss}).
Because $\hat{D}_T$ is itself a valid isovist, $\tilde{D}_T$ stays on or near the
geometry manifold, unlike Gaussian-blur corruption, which smooths across the sharp
depth discontinuities that real isovists always exhibit and therefore misrepresents
the inference-time error distribution.

\begin{algorithm}[t]
\caption{Self-Rollout Scheduled Sampling (per training step, curriculum weight $\lambda$)}
\label{alg:srss}
\begin{algorithmic}[1]
\Require Context $\{D_t\}_{t=1}^{T}$, actions $\{\mathbf{a}_t\}$, target $D_{T+1}$
\State \textbf{(no\_grad)} $\hat{D}_T \leftarrow
       f_\theta\!\left(\{D_t\}_{t=1}^{T-1},\, \mathbf{a}_{T-1\to T}\right)$
\State \textbf{(mix)} $\tilde{D}_T \leftarrow (1-\lambda)\,D_T + \lambda\,\hat{D}_T$
\State \textbf{(train)} $\hat{D}_{T+1} \leftarrow
       f_\theta\!\left(\{D_t\}_{t=1}^{T-1}\cup\{\tilde{D}_T\},\, \mathbf{a}_{T\to T+1}\right)$
\State $\mathcal{L} \leftarrow \mathcal{L}_{\text{depth}}(\hat{D}_{T+1}, D_{T+1})
       + 2.0\,\mathcal{L}_{\text{grad}}(\hat{D}_{T+1}, D_{T+1})$; update $\theta$ (AdamW)
\end{algorithmic}
\end{algorithm}

\begin{figure}[t]
  \centering
  \includegraphics[width=\linewidth]{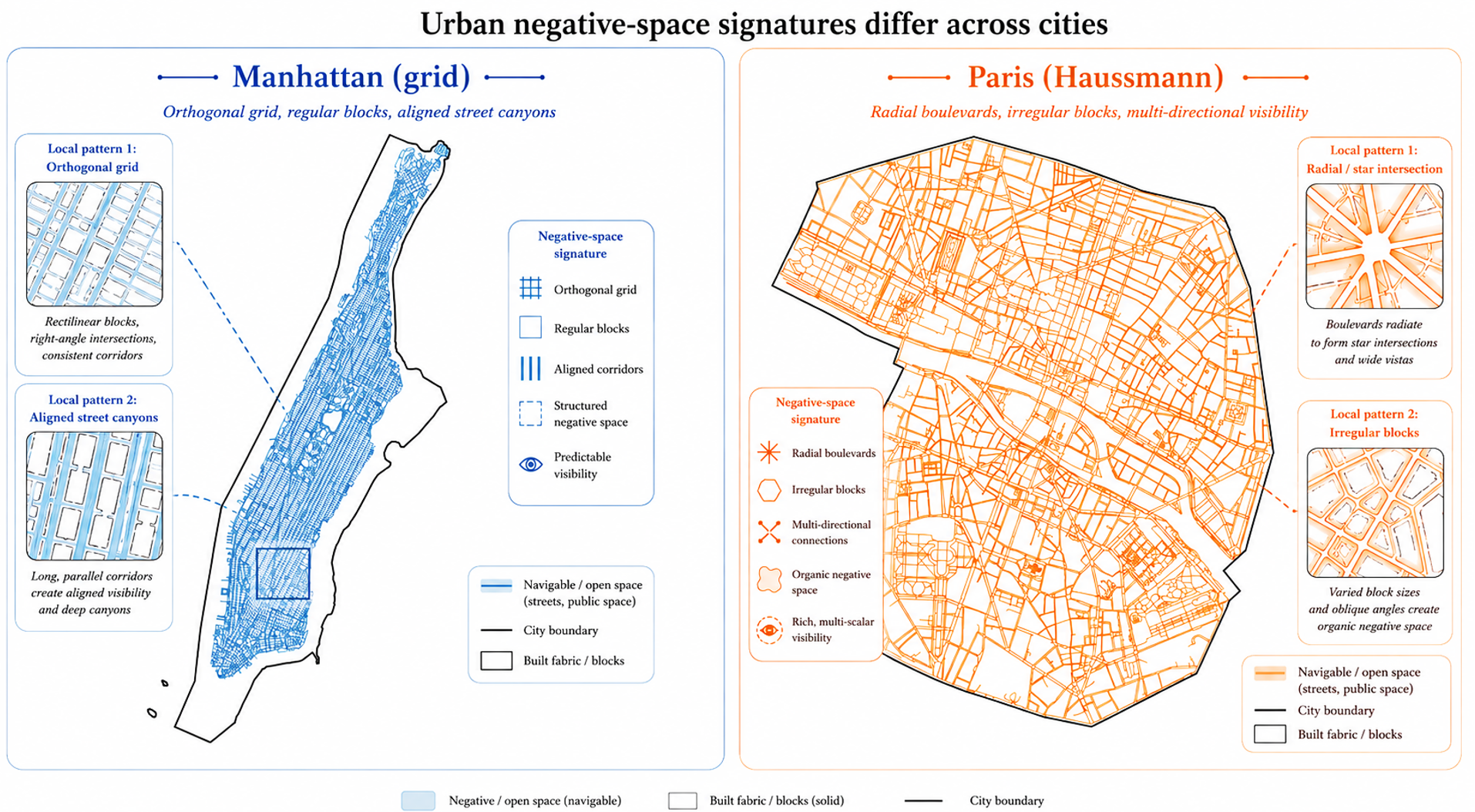}
  \caption{\textbf{Two morphologically distinct cities.} Walking road networks of
  Manhattan (gridded, $36{,}212$ nodes, $\approx$115\,km\textsuperscript{2}) and
  Paris arrondissements 1--6 (Haussmannian, $10{,}957$ nodes,
  $\approx$15.5\,km\textsuperscript{2}), both derived from OpenStreetMap. The two
  panels are not drawn to a common scale. The contrast in street geometry is the
  morphological basis of the cross-city signature.}
  \label{fig:dataset}
\end{figure}

\section{Experiment}
\label{sec:exp}

We evaluate three questions. (1) How accurate is single-step isovist prediction
against a strong baseline? (2) Does a city-blind world model develop a decodable
cross-city spatial signature, and is it explained by single-frame appearance?
(3) Does the persistent spatial map improve cross-path geometric consistency?
Single-step numbers are from the best checkpoint of a 400-epoch run; the signature
and consistency analyses are post-hoc and do not depend on that run. The two cities
under study are shown in Figure~\ref{fig:dataset}.

\subsection{Single-step prediction quality}
\label{sec:exp-single}

\begin{figure}[H]
  \centering
  \includegraphics[width=\linewidth]{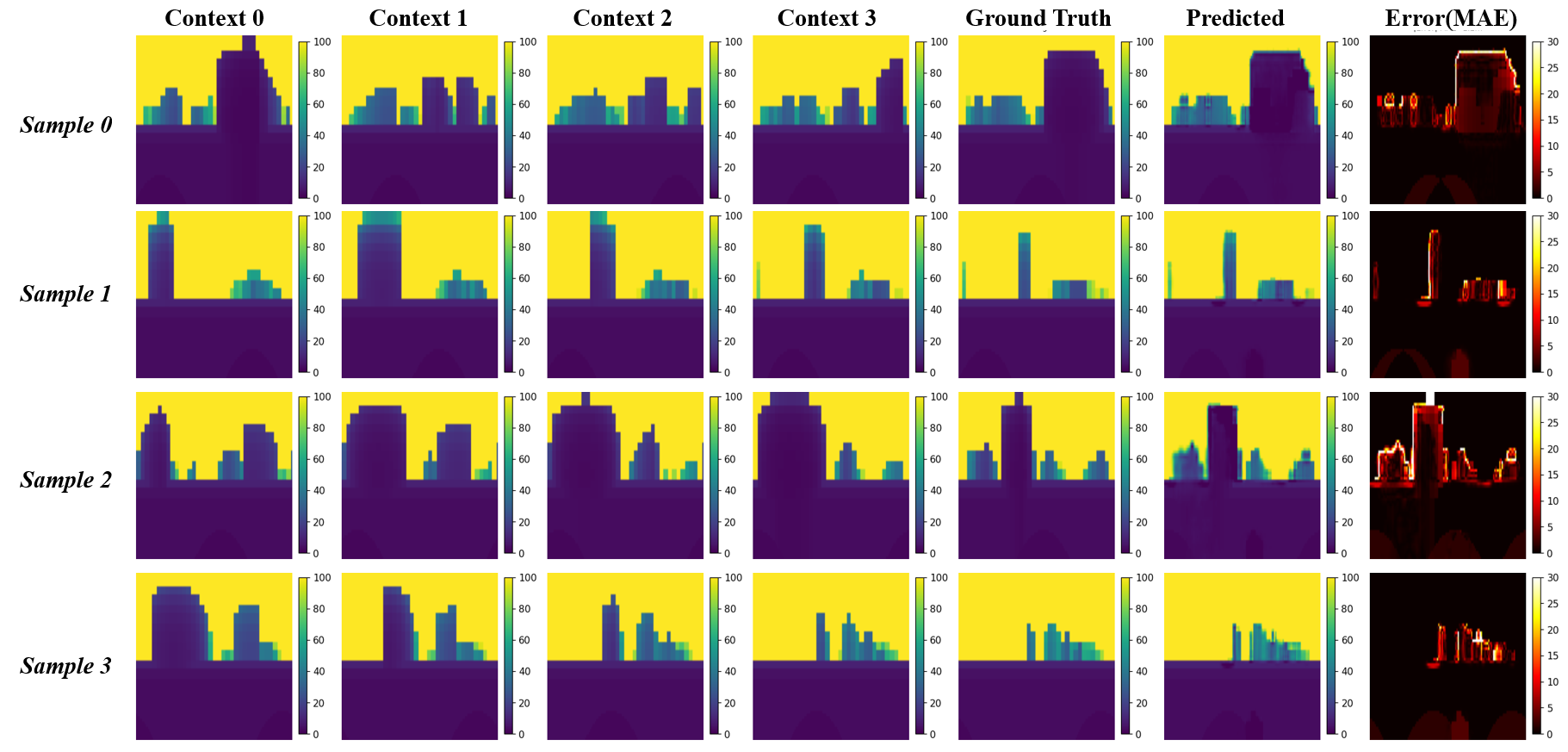}
  \caption{\textbf{Qualitative single-step prediction.} For four test samples we
  show (left to right) the four context isovists, the predicted next isovist, the
  ground truth, and the per-pixel absolute error. Depth in meters (viridis); error
  in meters (hot). Residual prediction preserves the sharp building edges inherited
  from the last context frame; errors concentrate at newly visible surfaces near the
  field-of-view edges.}
  \label{fig:qual}
\end{figure}

We compare against \textbf{copy-last}, the zero-parameter predictor
$\hat{D}_{T+1} = D_T$. At $10$\,m steps the isovist changes little between
consecutive frames, so copy-last is a \emph{strong} baseline; beating it on every
metric requires the model to improve precisely on the small, structured changes at
the field-of-view edges, which is the hardest part of the problem. We report MAE,
RMSE, SSIM, and building-edge F1 on the $n{=}832$ test samples
(Table~\ref{tab:single}) and show qualitative
context/prediction/ground-truth/error panels in Figure~\ref{fig:qual}. Our model
improves over copy-last on MAE ($3.57$ vs.\ $4.36$\,m), RMSE ($11.34$ vs.\
$13.80$\,m), and edge-F1 ($0.719$ vs.\ $0.689$), and matches it on SSIM ($0.852$
vs.\ $0.856$): it learns the small, structured inter-frame changes at the
field-of-view edges that copy-last cannot. Multi-step rollout results, where the
model continues to beat copy-last at every horizon, are reported in
Appendix~\ref{app:rollout} (Figure~\ref{fig:rollout}, Table~\ref{tab:multistep}).
The empirical centrepiece of the paper nonetheless remains the cross-city signature
rather than single-step state of the art.

\begin{table}[H]
  \caption{Single-step isovist prediction quality on the combined test set
  ($n{=}832$). MAE/RMSE in meters (relative to $R_{\max}=100$\,m); higher SSIM and
  Edge-F1 is better. Copy-last is a strong baseline because the isovist changes
  little over a $10$\,m step; our model (best checkpoint of a 400-epoch run) improves
  on it on every metric except SSIM, where the two are tied.}
  \label{tab:single}
  \centering
  \small
  \begin{tabular}{lcccc}
    \toprule
    Method & MAE (m)$\downarrow$ & RMSE (m)$\downarrow$ & SSIM$\uparrow$ & Edge-F1$\uparrow$ \\
    \midrule
    Copy-last (zero-parameter) & $4.36$ & $13.80$ & $\mathbf{0.856}$ & $0.689$ \\
    Ours (negative-space world model) & \maeOurs & \rmseOurs & \ssimOurs & \edgeOurs \\
    \bottomrule
  \end{tabular}
\end{table}

We also verify that city-blind training does not collapse onto one city: mean
validation loss is $0.184$ for Manhattan versus $0.290$ for Paris, a ratio of
$1.58\times < 2\times$. This margin supports the city-blind formulation; had one
city's loss exceeded the other's by more than $2\times$, an explicit
city-conditioned variant would have been warranted.

\subsection{The emergent cross-city spatial signature}
\label{sec:exp-signature}

\begin{figure}[t]
  \centering
  \includegraphics[width=\linewidth]{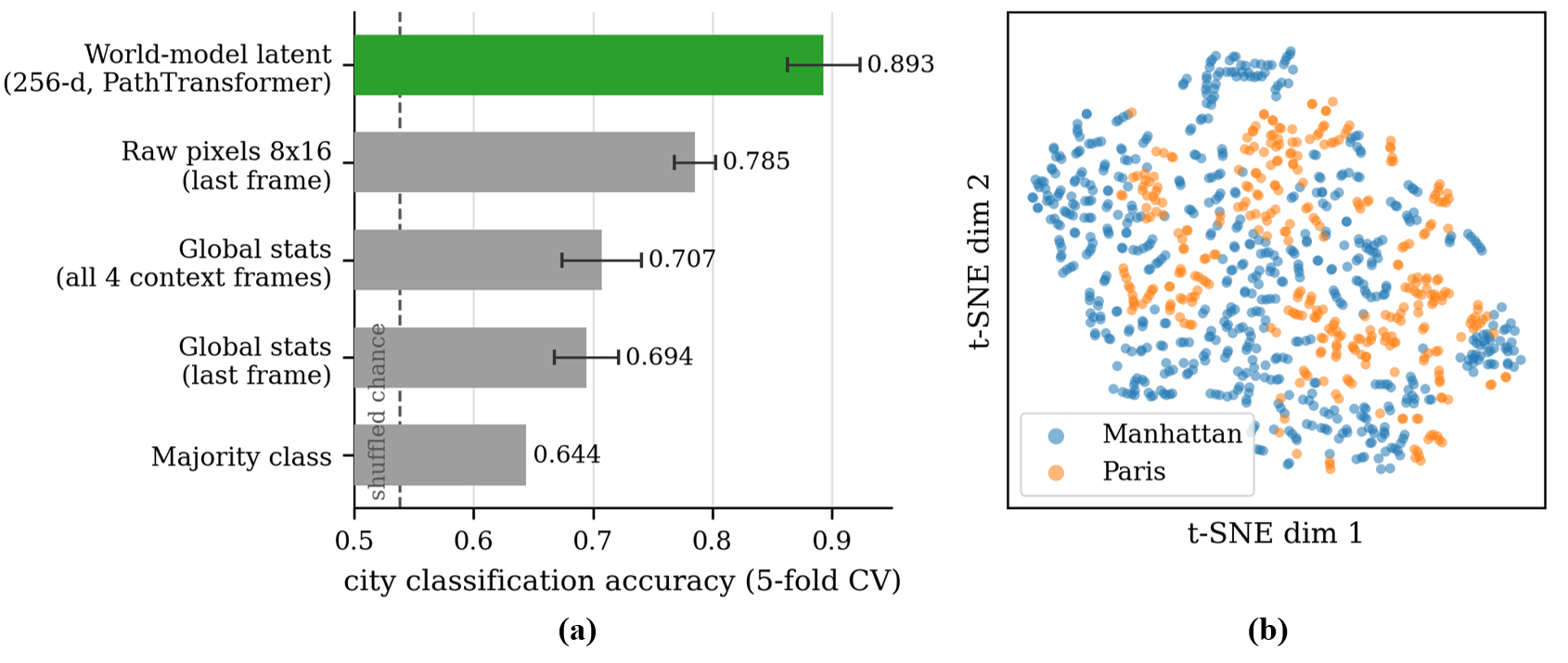}
  \caption{\textbf{Emergent cross-city spatial signature.} (a)~City-classification
  accuracy (five-fold CV) for a ladder of probes under an identical protocol. The
  city-blind world model's PathTransformer latent decodes the source city far above
  raw pixels, per-frame statistics, the majority class, and shuffled chance. A probe
  that beats single-frame baselines is reading structure absent from any single
  frame's appearance, in how the temporal model integrates the sequence and the
  action. (b)~t-SNE of the latents coloured by city; Manhattan and Paris occupy
  largely separable regions. (c)~Training convergence: train and validation loss over
  the 400-epoch run, with validation loss plateauing early.}
  \label{fig:signature}
\end{figure}

This is the paper's headline result. We test whether the single \emph{city-blind}
model has internalized city-distinctive structure by probing its latents. For every
validation sample we extract the $256$-d PathTransformer output at the last context
frame and train an $L_2$-regularized logistic regression to classify the source city
(Manhattan vs.\ Paris), reporting mean accuracy over stratified five-fold
cross-validation ($n{=}946$; $609$ Manhattan, $337$ Paris). To establish what the
signature is \emph{not}, we run the identical probe protocol on a ladder of baselines
(Table~\ref{tab:signature}, Figure~\ref{fig:signature}): the majority class;
shuffled labels (chance); single-frame global isovist statistics; the same
statistics over all four context frames; and the raw, $8\times16$-pooled pixels of
the last frame.

\begin{table}[H]
  \caption{\textbf{Cross-city signature baseline ladder} (five-fold CV, $n{=}946$:
  $609$ Manhattan / $337$ Paris). All rows use the identical probe protocol. The
  world-model temporal latent (bold) beats the strongest single-frame baseline (raw
  pixels) by $\sim$11 points and trivial frame statistics by $\sim$20, so the
  signature is \emph{not} explained by single-frame appearance.}
  \label{tab:signature}
  \centering
  \small
  \begin{tabular}{lc}
    \toprule
    Probe input & Accuracy (\%) \\
    \midrule
    Majority class & $64.4$ \\
    Shuffled labels (chance) & $53.8$ \\
    Single-frame global statistics & $69.4 \pm 2.7$ \\
    Global statistics over 4 context frames & $70.7 \pm 3.3$ \\
    Raw $8\times16$-pooled pixels (last frame) & $78.5 \pm 1.7$ \\
    \midrule
    \textbf{World-model PathTransformer latent (256-d)} & $\mathbf{89.3 \pm 3.0}$ \\
    \bottomrule
  \end{tabular}
\end{table}

The world-model latent reaches $89.3\% \pm 3.0\%$, exceeding the strongest
single-frame baseline (raw pooled pixels, $78.5\%$) by roughly $11$ points and the
single-frame statistics by roughly $20$. The crucial comparison is against raw
pixels and per-frame statistics: a probe that beats these is reading structure that
\emph{is not present} in any single frame's appearance. It lives in how the temporal
model integrates the sequence and the action. The signature is therefore a property
of the learned dynamics, not a trivially decodable surface feature. We emphasize the
probe uses a single city-blind model with \emph{no} city conditioning at any point
in training.

Training convergence is shown in Figure~\ref{fig:signature}(c): validation loss
plateaus early (by roughly epoch 30) while training loss continues to decrease, and
we report results from the best validation checkpoint. Per-epoch qualitative
evolution is shown in Appendix~\ref{app:dynamics}.

\begin{figure}[t]
  \centering
  \includegraphics[width=\linewidth]{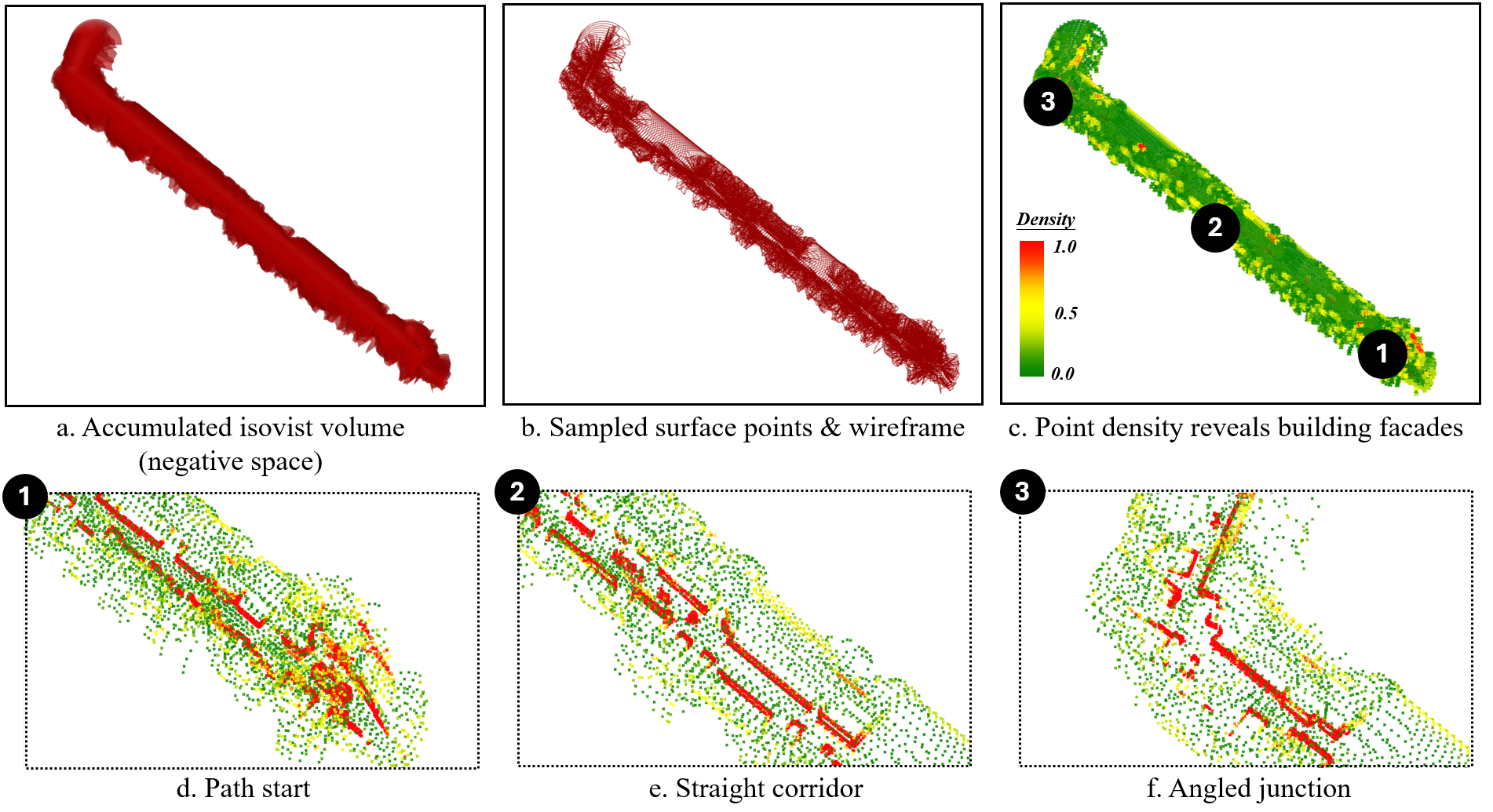}
  \caption{\textbf{Negative space gives rise to positive-space geometry.}
  Accumulating the world model's predicted isovists along a single path produces a
  continuous negative-space volume (a). Sampling its surface as points and wireframe
  (b), then colouring those points by local density (c), makes the bounding building
  facades emerge as high-density ridges (warmer $=$ higher density, i.e.\ more
  consistently reconstructed surface; green $=$ sparse open volume). The bottom row
  shows density-coloured detail views at three points along the path: the path start
  (d), a straight corridor (e), and an angled junction (f), where the facades
  recovered purely from accumulated visibility are clearest. This is the
  reconstructive property of the representation: the open volume an agent traverses,
  accumulated over a trajectory, recovers the positive-space surfaces that bound it.}
  \label{fig:recon}
\end{figure}

Finally, the negative-space representation is reconstructive: accumulating predicted
isovists along a path recovers the positive-space surfaces that bound the open
volume. Figure~\ref{fig:recon} illustrates this reconstructive property. As predicted
isovists accumulate along a path, their surface points trace out the surrounding
building facades, which appear as high-density ridges when the points are coloured by
local density.

\subsection{Spatial-map consistency ablation (preliminary)}
\label{sec:exp-ablation}

\begin{figure}[t]
  \centering
  \includegraphics[width=\linewidth]{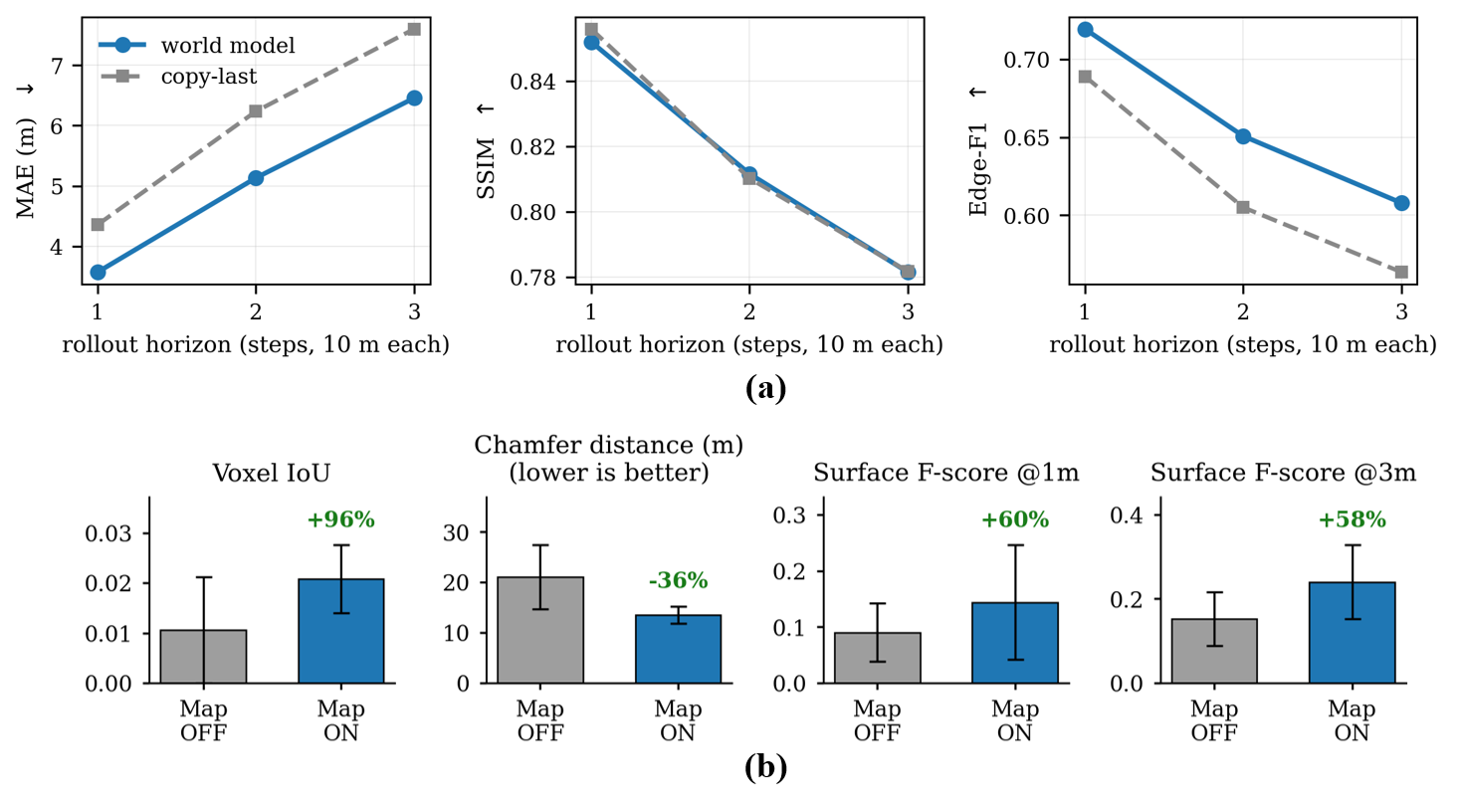}
  \caption{\textbf{Multi-step rollout and spatial-map consistency.} (a)~Prediction
  quality vs.\ rollout horizon for the world model and copy-last: the model beats
  copy-last on MAE and Edge-F1 at every horizon and tracks it on SSIM, with error
  growing gracefully rather than diverging. (b)~Spatial-map ablation (preliminary,
  synthetic): cross-path intersection consistency with the persistent BEV map ON vs.\
  OFF on a synthetic four-way crossing ($n{=}4$ pairs). The map improves all four
  metrics, but the large error bars at $n{=}4$ make this a directional
  proof-of-concept, not a validated result; real-city evaluation is future work.}
  \label{fig:rollout}
\end{figure}

Finally, we test whether the persistent spatial map improves cross-path geometric
consistency. We \emph{stress that this is a preliminary proof-of-concept}: it uses a
single synthetic four-way grid intersection with only $n{=}4$ crossing pairs and is
\emph{not} a validated consistency claim. We turn the persistent map ON vs.\ OFF and
measure agreement between the two crossing paths' accumulated point clouds inside the
intersection region (Table~\ref{tab:ablation}, Figure~\ref{fig:rollout}).

\begin{table}[t]
  \caption{\textbf{Spatial-map ON/OFF intersection consistency} ---
  \emph{preliminary proof-of-concept} on a synthetic four-way grid intersection,
  $n{=}4$ crossing pairs. Higher IoU/F-score and lower Chamfer is better. The map
  improves all four metrics; this is a directional signal, not a validated claim.}
  \label{tab:ablation}
  \centering
  \small
  \begin{tabular}{lcccc}
    \toprule
    Configuration & Voxel IoU$\uparrow$ & Chamfer (m)$\downarrow$
                  & F-score@1m$\uparrow$ & F-score@3m$\uparrow$ \\
    \midrule
    Spatial map OFF & $0.011$ & $21.1$ & $0.090$ & $0.152$ \\
    Spatial map ON  & $\mathbf{0.021}$ & $\mathbf{13.5}$ & $\mathbf{0.144}$ & $\mathbf{0.240}$ \\
    \midrule
    Relative change & $+91\%$ & $-36\%$ & $+60\%$ & $+58\%$ \\
    \bottomrule
  \end{tabular}
\end{table}

With the map on, all four metrics improve: voxel IoU $0.011\to0.021$ ($+91\%$),
symmetric Chamfer $21.1\to13.5$\,m ($-36\%$), F-score@1m $0.090\to0.144$ ($+60\%$),
and F-score@3m $0.152\to0.240$ ($+58\%$). We read this as encouraging evidence that
an explicit, writable BEV memory can reconcile geometry where paths cross. But with
only four synthetic pairs it is a proof-of-concept, not a validated result.
Establishing cross-path consistency on real Manhattan and Paris intersections at
scale, enabled by the released pipeline, is future work.

\section{Discussion}
\label{sec:disc}

\subsection{What the cross-city signature implies, and what it does not}
\label{sec:disc-implies}

A city-blind model that never receives a city label nonetheless encodes city
identity in its temporal latents at $89.3\%$, well above what any single frame's
appearance supports (raw pixels, $78.5\%$; frame statistics, $69.4\%$). The gap
between the temporal-latent probe and the single-frame probes is the load-bearing
comparison: it localizes the signature in the dynamics the model learns, how the
visibility volume evolves as the agent moves, rather than in static appearance.
Manhattan's regular grid and Paris's radial Haussmann blocks induce systematically
different isovist sequences (long straight corridors with abrupt intersection
openings versus oblique sightlines and angled junctions), and the model's latents
separate along that axis. This is the hidden geometry made legible: the model has
recovered, from movement alone, the morphological structure that space syntax has
long had to compute from a known plan. We want to be precise about the strength of
the claim, though. $89.3\%$ is linear decodability on a binary problem, which
establishes that a city-distinctive axis exists in the representation, not that the
model has built an explicit map, performed localization, or ``understood''
morphology in any richer sense. The honest reading is that predictive training on
negative space yields representations from which coarse urban structure is
recoverable, unprompted; the richer interpretations are hypotheses the present
experiment does not test.

\subsection{Relation to emergent cognitive maps}
\label{sec:disc-cognitive}

This phenomenon sits in a clear lineage. \citet{gornet2024cognitive} show that a
visual predictive-coding network trained to predict the next view in a small, fixed
environment develops an emergent, implicit cognitive map whose latents support
localization. Our cross-city signature is a real-city instance of the same premise,
that predictive training yields spatially meaningful representations, observed at the
coarser level of city morphology rather than within-environment position. The
connection to neuroscience-inspired models is by analogy and should be read as such:
grid-like and place-like codes emerge in artificial agents trained on navigation
objectives~\citep{banino2018grid, cueva2018emergence}, and we observe an analogous
\emph{unsupervised} emergence of a city-level code without any localization or
city-classification objective in the loss. We draw this analogy deliberately
narrowly. We do not claim a grid-cell mechanism, a metric place code, or any
correspondence to neural representation beyond the shared high-level observation that
a navigation-prediction objective is sufficient to induce decodable spatial
structure. The two architectural differences from Gornet and Thomson remain the
substantive ones: our spatial map is explicit, persistent, and writable rather than
implicit and read out post hoc, and our prediction target is geometric negative space
rather than appearance.

\subsection{What the negative-space representation reveals}
\label{sec:disc-reveals}

It is worth stating plainly what this paper does and does not reveal, since that is
its central claim. Across both single-step prediction and the emergent signature, the
common thread is that negative space exposes geometry that appearance hides. A
$64\times128$ spherical isovist carries no lighting, shadow, or colour, only the
metric distance to surrounding surface in every direction, so what the model learns
to predict is, by construction, the navigable shape of the space rather than its
photometric skin. Two consequences follow. First, at the level of a single step, the
structure the model must get right is the field-of-view geometry where new surfaces
appear and old ones occlude. This is why it improves over copy-last precisely on
edge-F1, the most geometry-sensitive metric. Second, at the level of a whole
trajectory, the accumulated visibility volume recovers the positive-space surfaces
that bound it, and the way that volume evolves under movement differs systematically
between a grid and a radial plan. This is what the city-blind probe reads out. In
both cases the model is recovering the hidden, navigable geometry that supports how
an embodied agent moves on the ground and through above-ground form. This is the
structure space syntax has historically had to compute from a known plan, here
learned from movement alone.

\subsection{Why residual prediction and self-rollout SS help}
\label{sec:disc-why}

For $\sim$10\,m steps the dominant signal between consecutive isovists is shared: the
same facades, shifted slightly in azimuth. Residual prediction turns the task into a
sparse update concentrated on genuinely changing regions, letting the decoder inherit
sharp boundaries rather than re-synthesize them. This is why we beat copy-last
precisely on edge-F1, the metric most sensitive to the small structured changes at
the field-of-view edges. Self-rollout corruption, unlike Gaussian blur, produces the
very error types the model meets at inference, such as small metric offsets on flat
surfaces and occasional mis-predictions in partially occluded regions, because the
corruption is the model's own valid isovist and so lies on the geometry manifold
rather than off it.

\subsection{Limitations and future work}
\label{sec:disc-limits}

Several constraints bound our claims. \emph{(1) Two cities.} The signature is
demonstrated on two morphologically distinct cities (grid vs.\ Haussmann); a third,
morphologically distinct city is the obvious next step and is enabled by the released
pipeline. \emph{(2) Binary decodability.} $89.3\%$ is linear separability on a
two-class problem; it establishes a decodable axis, not an explicit or metric map.
\emph{(3) Single-step scope.} Our 400-epoch model beats copy-last on
MAE/RMSE/edge-F1 and matches it on SSIM, but we do not claim single-step SOTA against
external depth-prediction methods, which target a different, RGB-conditioned setting.
\emph{(4) Consistency is preliminary.} The spatial-map ablation uses one synthetic
intersection with $n{=}4$ pairs and is explicitly not a validated consistency result;
a real-city, multi-intersection evaluation is future work. \emph{(5) Lossy
representation.} A $64\times128$ depth map captures only the nearest surface per
direction, so a single isovist is a single-layer spherical shell; multi-frame fusion
only partially compensates. \emph{(6) Height provenance.} $34.2\%$ of Paris heights
are neighbour-median imputed; some of the cross-city signature could reflect this
provenance difference rather than pure morphology, a confound we disclose rather than
hide. \emph{(7) Static, synthetic geometry.} Dynamic objects and the sim-to-real gap
to real range sensors remain unaddressed.

The released pipeline scales the signature study to many cities and supports a proper
multi-city consistency evaluation; with three or more morphologically distinct cities
the binary-decodability limitation becomes a multi-class one, and morphology becomes
a learnable axis rather than a single separating hyperplane. Real LiDAR spherical
range images are a direct drop-in for ray-cast isovists, and appearance or semantics
can be added as extra channels of the spherical representation without changing the
predictive architecture. This turns the photometric variation we currently exclude
into optional conditioning rather than entangled input.

\section{Conclusion}
\label{sec:conc}

We introduced a world model that operates in \emph{negative space}. It predicts the
agent's 3D isovist, the spherical visibility volume between buildings, from a short
history and a movement action, using residual depth prediction to inherit sharp
building edges and self-rollout scheduled sampling to close the teacher-forcing gap
on the geometry manifold. To enforce cross-path consistency we equipped the model
with a persistent, explicit, writable BEV latent spatial map. Our central empirical
finding is that a single city-blind model trained on Manhattan and Paris develops an
emergent cross-city spatial signature: city identity is linearly decodable from its
temporal latents at $89.3\%$, beating raw-pixel and single-frame-statistic probes and
thereby ruling out single-frame appearance as the explanation. Our predictor improves
over a strong copy-last baseline on MAE, RMSE, and edge sharpness (matching it on
SSIM), we give a preliminary proof-of-concept that the spatial map improves
intersection consistency, and we release a reproducible two-city OpenStreetMap isovist
dataset and pipeline. We do not overclaim: the signature rests on two cities, and the
consistency result is synthetic and preliminary. What we do claim is that
negative-space world modelling is a lightweight, interpretable, and reproducible
foundation for urban spatial reasoning, one that gives rise, unprompted, to structure
that knows which city it is walking through.

\section*{Data and Code Availability}

We release a reproducible isovist-sequence dataset for Manhattan and Paris. All
source geometry derives from OpenStreetMap~\citep{haklay2008osm, openstreetmap};
accordingly, the dataset is distributed under the \textbf{Open Database License
(ODbL)}. We provide a datasheet following \citet{gebru2021datasheets}, including
per-city height-provenance tables that disclose the substantial Paris height
imputation. The dataset will be archived on Zenodo under ODbL. The full generation
pipeline and model code will be released publicly upon acceptance.

\bibliographystyle{plainnat}
\bibliography{references}

\clearpage
\appendix
\setcounter{figure}{0}
\setcounter{table}{0}
\renewcommand{\thefigure}{A\arabic{figure}}
\renewcommand{\thetable}{A\arabic{table}}

\section{Additional experimental results}
\label{app:results}

\subsection{Multi-step autoregressive rollout}
\label{app:rollout}

Beyond the single-step results of Section~\ref{sec:exp-single}, we evaluate the world
model under autoregressive rollout, feeding its own predictions back as context. We
report prediction quality at horizons of 1, 2, and 3 steps (10\,m each) in
Table~\ref{tab:multistep}, and visualize the corresponding trends in
Figure~\ref{fig:rollout}(a). The world model beats copy-last on MAE, RMSE, and
Edge-F1 at every horizon and tracks it on SSIM. Error grows gracefully rather than
diverging, consistent with the self-rollout scheduled-sampling training.

\begin{table}[H]
  \caption{Multi-step rollout metrics (combined test set). MAE/RMSE in meters; higher
  SSIM and Edge-F1 is better. World model vs.\ copy-last at each horizon.}
  \label{tab:multistep}
  \centering
  \small
  \begin{tabular}{lcccc}
    \toprule
    Horizon & MAE (m) model/copy & RMSE (m) model/copy & SSIM model/copy & Edge-F1 model/copy \\
    \midrule
    1 (10\,m) & $3.57 / 4.36$ & $11.34 / 13.80$ & $0.852 / 0.856$ & $0.719 / 0.689$ \\
    2 (20\,m) & $5.13 / 6.24$ & $14.40 / 17.45$ & $0.812 / 0.810$ & $0.651 / 0.605$ \\
    3 (30\,m) & $6.46 / 7.60$ & $16.78 / 19.72$ & $0.781 / 0.782$ & $0.608 / 0.564$ \\
    \bottomrule
  \end{tabular}
\end{table}

\subsection{Training dynamics}
\label{app:dynamics}

\begin{figure}[H]
  \centering
  \includegraphics[width=0.7\linewidth]{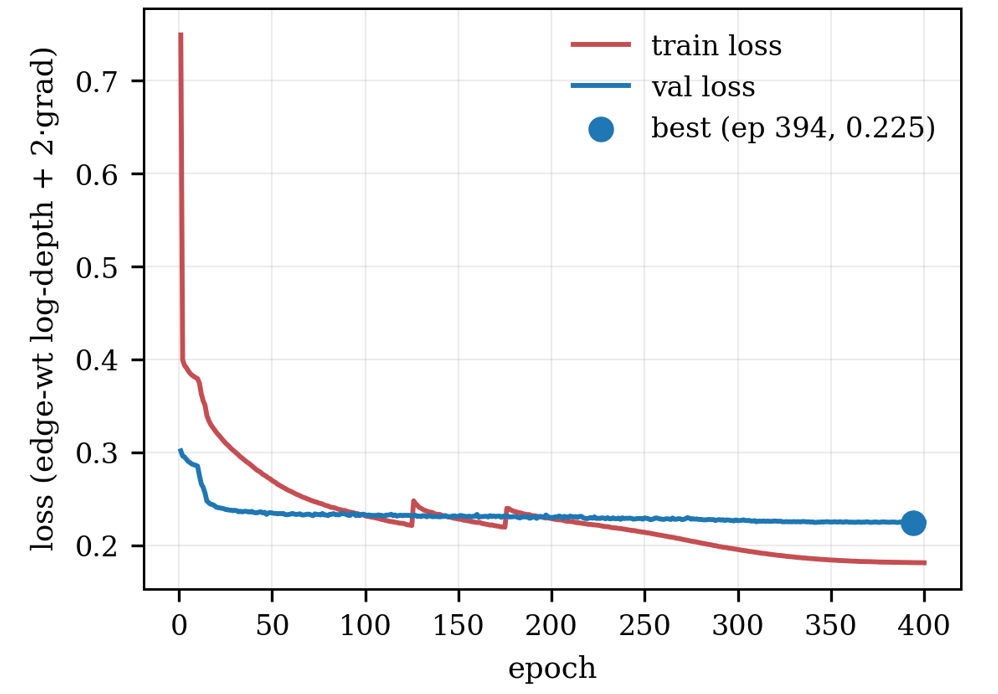}
  \caption{\textbf{Training convergence.} Train and validation loss over 400 epochs;
  validation loss plateaus early while training loss keeps decreasing, with the best
  validation checkpoint marked.}
  \label{fig:convergence}
\end{figure}

\begin{figure}[H]
  \centering
  \includegraphics[width=\linewidth]{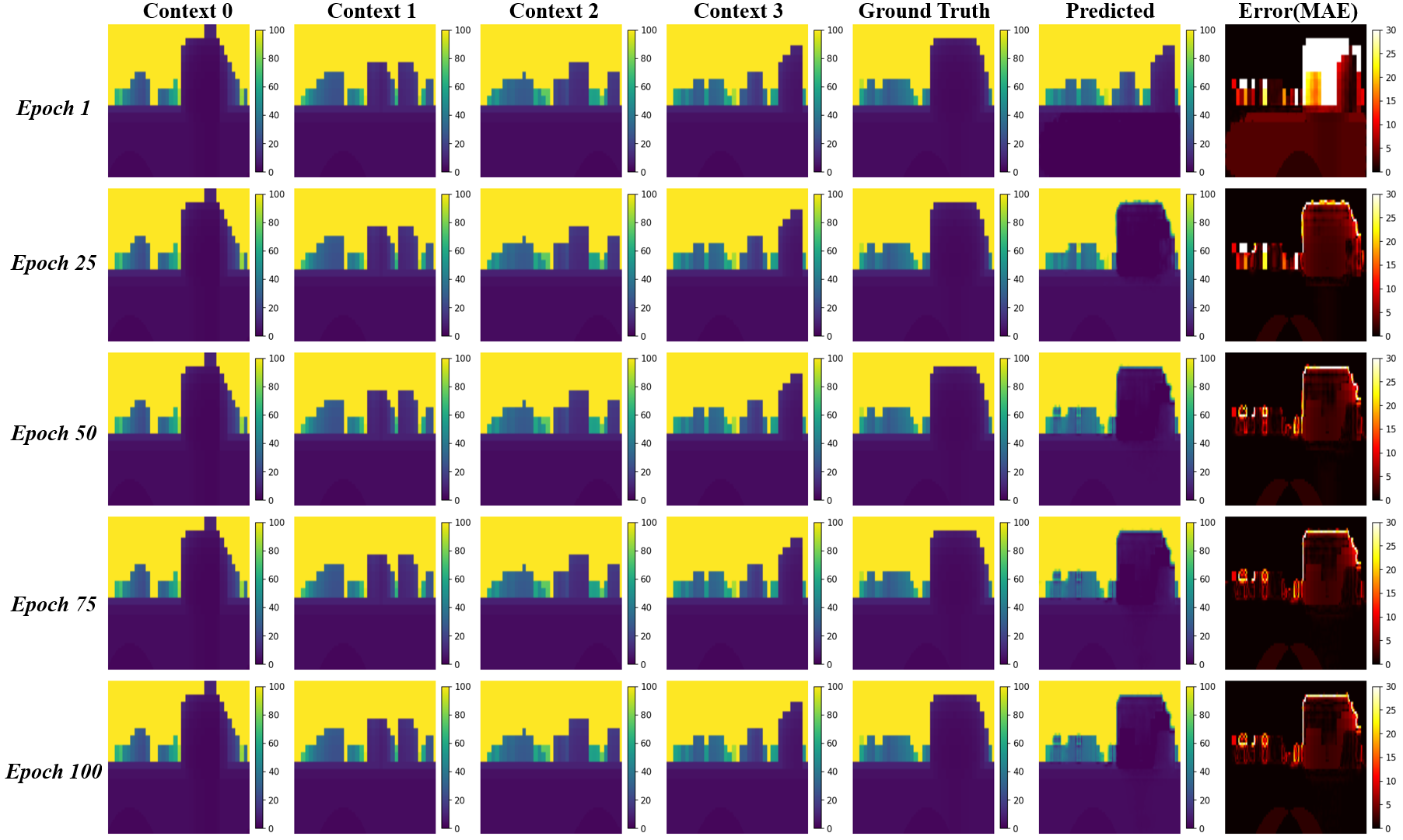}
  \caption{\textbf{Per-epoch evolution of single-step prediction.} The same sample
  predicted at epochs 1, 25, 50, 75, and 100. The error contracts to a thin edge
  contour as training proceeds.}
  \label{fig:perepoch}
\end{figure}

We characterize training along two complementary axes. Figure~\ref{fig:convergence}
shows convergence over the full 400-epoch run: validation loss plateaus early, by
roughly epoch 30, while training loss continues to decrease, and we report results
from the best validation checkpoint (epoch 394). Figure~\ref{fig:perepoch} then shows
how single-step prediction improves visually over the same run, with the same sample
predicted at epochs 1, 25, 50, 75, and 100. Early predictions miss the emerging
surface (large bright error); by roughly epoch 50 the model recovers sharp building
edges and the error collapses to a thin contour at the field-of-view boundary,
consistent with the early plateau in validation loss.

\section{Interactive viewer}
\label{app:viewer}

\begin{figure}[H]
  \centering
  \includegraphics[width=\linewidth]{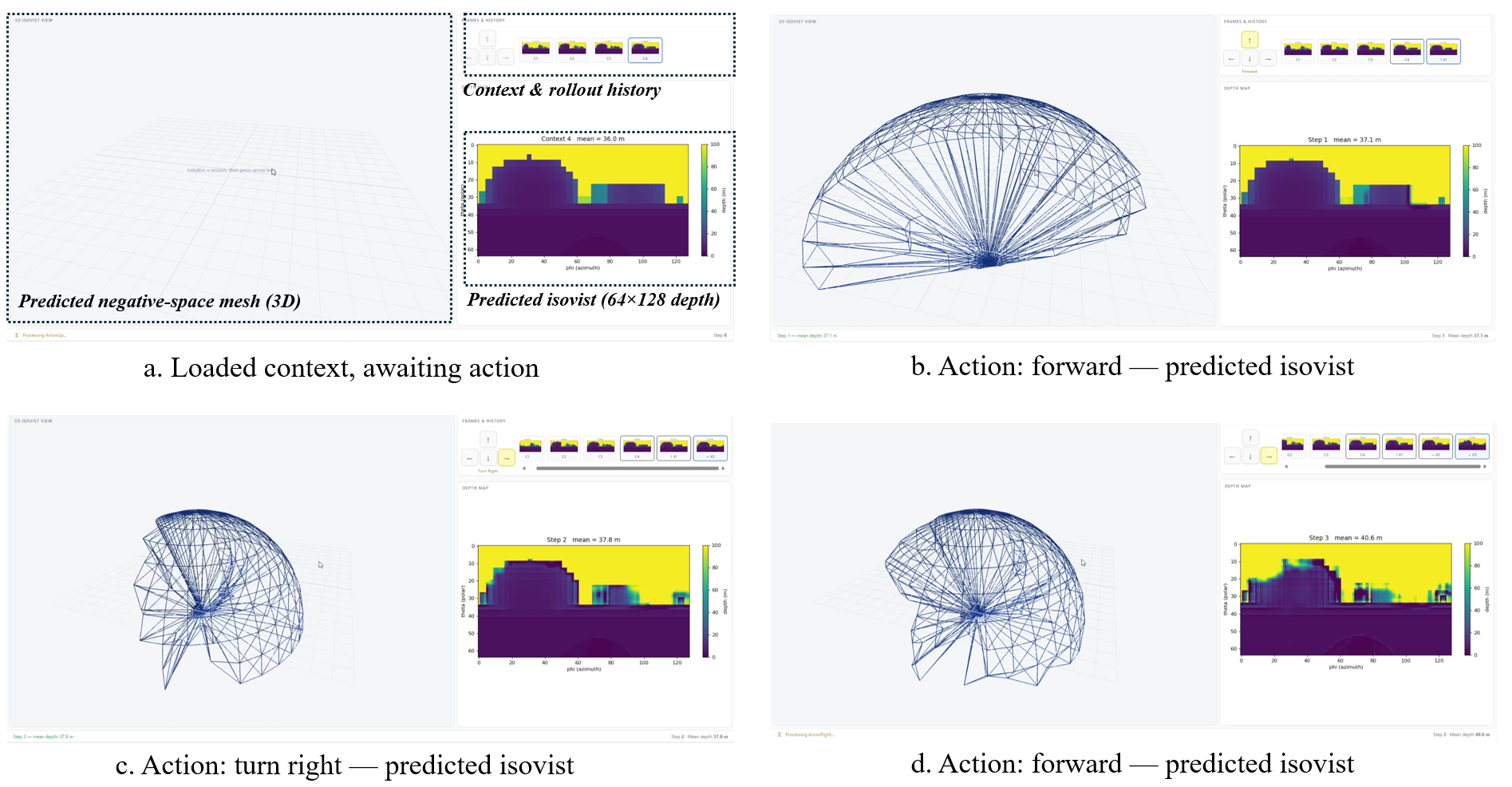}
  \caption{\textbf{Interactive isovist rollout.} A browser-based viewer drives the
  world model with keyboard actions. After loading a path (1), each arrow-key action
  generates the next isovist autoregressively: forward $\to$ Step 1 (2), turn right
  $\to$ Step 2 (3), forward $\to$ Step 3 (4). Each panel shows the predicted
  negative-space hemisphere (left) and its $64\times128$ depth map (right); mean
  depth is annotated per step.}
  \label{fig:viewer}
\end{figure}

We release a browser-based interactive viewer that lets a user drive the world model
with keyboard actions. After loading a path, each arrow-key action generates the next
isovist autoregressively, and the predicted negative-space hemisphere and its
$64\times128$ depth map update in real time (Figure~\ref{fig:viewer}). The viewer is
intended as a qualitative demonstration of action-conditioned generation and is
included with the released code.

\end{document}